# Research on Multi-hop Inference Optimization of LLM Based on MQUAKE Framework

**Zucheng Liang[1,4], Wenxin Wei[1,5], Kaijie Zhang[2,6], Hongyi Chen[3,7]**

[1] China University of Mining and Technology, Jiangsu, China
[2] College of Engineering, Northeastern University, Boston, MA, USA
[3] University of California, Los Angeles, CA, USA

[4]1812503968@qq.com
[5]1527063371@qq.com
[6]zhang.kaij@northeastern.edu
[7]henrychyy@g.ucla.edu

**Abstract.** Accurately answering complex questions has consistently been a significant challenge for Large Language Models (LLMs). To address this, this paper proposes a multi-hop question decomposition method for complex questions, building upon research within the MQUAKE framework. Utilizing the LLAMA3 model, we systematically investigate the impact of multi-hop question decomposition within knowledge graphs on model comprehension and reasoning accuracy, both before and after model training. In our experiments, we systematically partitioned and converted the MQUAKE-T dataset into two distinct formats: a single-hop dataset designed for directly answering complex questions, and a multi-hop dataset constructed using the multi-hop question decomposition method. We then fine-tuned the LLAMA3 model on these datasets and conducted inference tests. Our results demonstrate that, without fine-tuning the LLM, the prediction performance based on the multi-hop question decomposition method significantly outperforms the method of directly answering complex questions. After fine-tuning using the LoRA (Low-Rank Adaptation) method, the performance of both approaches improved compared to the untrained baseline. Crucially, the method utilizing multi-hop decomposition consistently maintained its superiority. These findings validate the effectiveness of the multi-hop decomposition method both before and after training, demonstrating its capability to effectively enhance the LLM's ability to answer complex questions.

## 1. Introduction

In recent years, Large Language Models (LLMs) have achieved remarkable progress in Natural Language Processing (NLP), excelling particularly in text generation and comprehension tasks. Despite these advancements, LLMs still face significant challenges in handling complex tasks that require multi-step logical reasoning. Multi-hop reasoning demands iterative cycles of information retrieval and inference, which place higher requirements on knowledge representation, logical reasoning, and the ability to model long-range contextual dependencies [1,2].

Knowledge Graphs (KGs), as structured forms of knowledge representation, interconnect vast background information through (entity, relation, entity) triplets, offering rich retrievable sources for multi-hop reasoning [3]. However, most existing research has concentrated on single-hop reasoning, with limited effective approaches to multi-hop scenarios [4]. Current limitations lie in maintaining continuity and consistency across reasoning chains and in adapting to long-distance dependencies and information fusion inherent in multi-hop tasks.

Domestic research has made some progress in addressing these challenges. Chen Mengke introduced WH-CoT, a Chain-of-Thought prompting framework based on the 6W2H principle, to enhance LLM reasoning through structured prompting [5]. While promising, its generalization and practical efficiency require further validation. Xing Cao and Yun Liu proposed a novel model architecture for multi-hop question answering, combining coarse-grained question decomposition (CGDe) with fine-grained interaction (FGIn), demonstrating superior performance over state-of-the-art baselines on SQuAD and HotpotQA datasets [6]. Nevertheless, its adaptability to broader application contexts remains uncertain.

Internationally, the GPT-4 Technical Report by Achiam et al. emphasized progress in reasoning, hallucination mitigation, and interactivity across multi-task, multilingual, and multimodal evaluations [7]. Yet GPT-4 continues to show limitations in complex tabular reasoning. Deng et al. further proposed an approach integrating Abstract Meaning Representation (AMR) with text generation models for multi-hop question answering, trained on the QDMR dataset [8]. This method effectively decomposes complex questions but faces challenges in real-world deployment.

To address these issues, this paper introduces a multi-hop question decomposition strategy designed to improve reasoning effectiveness by splitting original complex questions into manageable sub-questions [9]. Building on Mello's decomposition method, experiments are conducted using the LLAMA3 large language model to validate the feasibility and effectiveness of the proposed approach in tackling complex multi-hop reasoning tasks.

## 2. Related Work

*2.1 Applications of Various Models in Knowledge Graphs*
In recent years, knowledge graphs (KGs) have been widely applied in information retrieval, question answering systems, and knowledge reasoning due to their unique advantages in information organization and knowledge representation[10]. For instance, by modeling structured knowledge, KGs can enhance semantic understanding and reasoning capabilities in natural language processing tasks. However, traditional pre-trained language models, which primarily rely on large-scale open-domain corpora for pre-training, exhibit significant limitations when handling multi-hop reasoning and domain-specific tasks. For example, BERT underperforms on complex tasks involving domain knowledge. To address this limitation, K-BERT attempted to improve performance on domain-specific tasks by incorporating domain knowledge graph information, thereby demonstrating the enhancing effect of KGs on language models. Nevertheless, due to BERT's relatively small parameter size and weak text generation capability, the improvement from KG integration remains limited.

In contrast, large language models (LLMs) such as LLAMA3, with their powerful generative capabilities, demonstrate significant potential when combined with KGs for complex problem-solving. These models are not only suitable for traditional question-answering tasks but can also efficiently process multi-hop relational chains through complex question decomposition. In complex QA tasks, LLAMA3's superior text feature extraction and generation capabilities, enabled by its massive parameter size, allow it to more effectively address dynamic knowledge updates and complex reasoning requirements, achieving higher performance ceilings compared to traditional methods.

*2.2 Knowledge Graph-Based Approaches for Complex Problem Solving: Methods and Limitations*
When addressing complex problems, researchers typically require deep modeling of multi-hop relationships. However, many traditional approaches focus on single-hop reasoning, failing to fully leverage rich hierarchical structures and contextual information for knowledge graph construction. These methods often rely on simplistic feature extraction techniques to perform prediction tasks. In multi-step reasoning scenarios, such approaches tend to lose critical semantic and structural information.

In recent years, the integration of large language models (LLMs) with knowledge graph analysis has opened new pathways for tackling complex questions. Inspired by knowledge graph principles, these methods transform data into natural language prompts, implicitly capturing deep entity relationships in latent spaces and significantly enhancing the understanding of complex relational patterns. For instance, the KG-LLM framework combines instruction fine-tuning with Chain-of-Thought (CoT) prompting to improve multi-hop link prediction performance, demonstrating LLMs' potential in handling multi-step reasoning tasks. Nevertheless, existing methods still exhibit notable limitations. First, many models prioritize discriminative prediction while lacking interpretability in generative reasoning. Second, they often lack flexible knowledge-editing capabilities, hindering timely adaptation to dynamic knowledge updates. Although studies show that models combining In-Context Learning (ICL) and CoT techniques can partially mitigate these issues, enhancing generalization performance on unseen tasks remains a critical unsolved challenge.

*2.3 Multi-hop Decomposition: Definition, Research Advances, and Challenges*

Multi-hop decomposition refers to a methodological approach that dissects complex problems into a series of interconnected sub-questions, enabling stepwise reasoning to reduce cognitive complexity and enhance accuracy. A representative example in multi-hop question answering would be processing the query: "Which authors won the Nobel Prize in Literature after receiving an Oscar?" This necessitates two distinct reasoning steps: first identifying all Oscar-winning authors, then determining which of these subsequently received the Nobel Prize in Literature.

**Key Challenges in Practical Implementation:**

  1. **Intermediate Step Generation**

Current methodologies predominantly employ rule-based or heuristic approaches to identify intermediate reasoning nodes through semantic relationship analysis between key entities. These methods demonstrate significant limitations when handling questions featuring:

- Highly complex or variable structural patterns.
- Substantial implicit information requirements.
- Context-dependent logical relationships.

  2. **Reasoning Chain Consistency Maintenance**

The reliability of multi-hop reasoning critically depends on maintaining strict logical coherence throughout the entire reasoning process. However, existing approaches frequently exhibit:

- Over-reliance on superficial pattern matching rather than deep logical inference.
- Inconsistent intermediate conclusions that propagate through the reasoning chain.
- Limited capacity for cross-domain knowledge integration.

*2.4 MQUAKE Methodology Analysis*

The MQUAKE framework specifically targets multi-hop reasoning tasks by dynamically adapting original knowledge graphs to optimize large language models (LLMs) for complex reasoning scenarios. Its core objective centers on enhancing the multi-hop reasoning capabilities of LLMs.

This framework incorporates two principal datasets: MQUAKE-CF and MQUAKE-T. MQUAKE-CF evaluates models' multi-hop reasoning performance under counterfactual edits by introducing hypothetical factual modifications. Conversely, MQUAKE-T assesses model adaptability to dynamic knowledge updates based on real-world temporal changes.

For instance, MQUAKE-T simulates events such as "the transition of UK Prime Minister from Boris Johnson to Rishi Sunak," requiring models to resolve derivative multi-hop questions after knowledge edits. Such tasks demand that models correctly trace and logically combine interrelated facts to generate consistent answers. Through these designs, the MQUAKE framework not only evaluates model performance on single-fact modifications but more critically examines their cross-fact integration capabilities during multi-hop reasoning. This provides essential analytical foundations for addressing multi-step reasoning challenges in complex scenarios.

**3. Research Design**

This study focuses on model training through LLaMA-Factory and converting the MQuAKE-T dataset into a format acceptable to LLaMA-Factory for training and evaluation in multi-hop inference tasks.

*3.1 Data set format transformation and slicing*

Since LLaMA-Factory currently only supports datasets in Alpaca format and ShareGPT format, we need to convert the original MQuAKE-T dataset into a format that is compatible with LLaMA-Factory.The MQuAKE-T dataset contains other redundant information. Therefore, it cannot be used directly for input, and we transformed it into a JSON structure that conforms to the Alpaca format in order to support multi-hop inference.

During the data format transformation process, we first specified the core fields of each data item, which mainly include:

- INSTRUCTION: indicates the core requirement of the problem, the PROMPT given to LLM.
- INPUT: problem with multiple decomposition chains.
- output: the final answer.
- history: record the intermediate questions and answers in the multi-hop reasoning process.

**The transformation process is as follows:**

First perform data parsing and extraction . Scripts are written to parse the raw MQuAKE-T data, extract the multi-hop fact chains, question text, and final answers, and clean and de-duplicate the data to ensure that key fields are retained intact. Then perform field mapping . The task requirements and examples are first entered into INSTRUCTION, the original question and multi-hop chain are mapped to the INPUT field, and the final answer is mapped to the OUTPUT, encapsulated as a single element. Then all the elements are stored in JSON file.

In order to validate the effectiveness of the multi-hop decomposition method, we process the dataset into two categories for comparison experiments. They are multi-hop dataset and single-hop dataset. Based on the transformed data in the previous section, we process. The multi-hop dataset is directly sliced from the transformed data, of which 70% is used as the training set and 30% as the testing set. The single-hop dataset removes multiple chains from the transformed data, i.e., the dataset is transformed into a direct answer to a complex question, and is sliced in the same way as the multi-hop dataset. This synchronized slicing ensures that the performance difference between multi-hop and single-hop tasks can be accurately observed during the training and evaluation phases.

*3.2 Model Training*

In this study, we work on the pre-trained LLAMA3 model,which possesses strong text generation and logical reasoning capabilities and is suitable for multi-hop reasoning tasks. We perform fine-tuning on it to verify the generalizability of the multi-hop problem decomposition approach for accuracy improvement. The fine-tuning method uses LoRA (Low-Rank Adaptation) technique to reduce memory and computational costs by updating only the parameters of the low-rank matrix layer, while accelerating the training process for efficient parameter updating and resource utilization. The training set is the part of the training set of the single-hop and multi-hop datasets cut in the previous section, and the two datasets are fed into the model separately. The two datasets are inputted into the model separately. Subsequently, the two models are trained based on the single-hop and multi-hop training sets according to the parameters in the YAML configuration file.

The core parameters for training were set as follows:

Table 1 Important Parameters

| Parameter name | Value | Description |
| --- | --- | --- |
| per_device_train_batch_size | 1 | Training batch size for each device, 1 sample processed at a time |
| gradient_accumulation_steps | 8 | The number of gradient accumulation steps, which corresponds to a total batch size of 8. |
| learning_rate | 1.0e-4 | The learning rate is set to 0.0001 to control the speed of parameter update. |
| num_train_epochs | 2 / 10 | Number of training cycles, the number of times the model traverses the training dataset. |

*3.3 Evaluation and comparison*

In order to evaluate the effect of multi-hop decomposition on accuracy in multiple scenarios, we use both untrained and trained LLAMA3 models to perform inference on multi-hop and single-hop test sets, respectively, by comparing the accuracy of the two datasets in the validation set at different stages of training for both the multi-hop and single-hop test sets. The accuracy improvement of the multi-hop question decomposition method over direct answers is evaluated against the effect of maintaining the advantage of the multi-hop decomposition method after model training.

To measure the performance of the model on the inference task, we adopt the following evaluation criterion: a word predicted by the model is considered to be correct if it agrees with the original label or its synonyms (alias word list). This evaluation strategy reduces the reliance on single-answer expressions and is more relevant to the diverse output scenarios of natural language. For example, when the original label is "Gautama Buddha", if the model predicts "Siddhartha Gautama", it can still be judged as correct through alias matching. After this determination, the number of correct and incorrect predictions is counted, and the accuracy formula in the sklearning library is called to calculate the result directly.

## 4. Experimental Result

The following table shows the accuracy of the model on the single-hop and multi-hop tasks for different experimental configurations, and the corresponding accuracy improvement:

Table 2 Comparison of results of each model

| Experimental Configuration | Single Jump Accuracy | Multi-hop accuracy | Accuracy improvement (‰) |
| --- | --- | --- | --- |
| Not fine-tuned (base) | 25.47% | 25.93% | 4.67 |
| LoRA Generation 2 Training | 88.89% | 89.32% | 4.33 |
| LoRA Generation 10 Training | 90.33% | 90.44% | 1.11 |

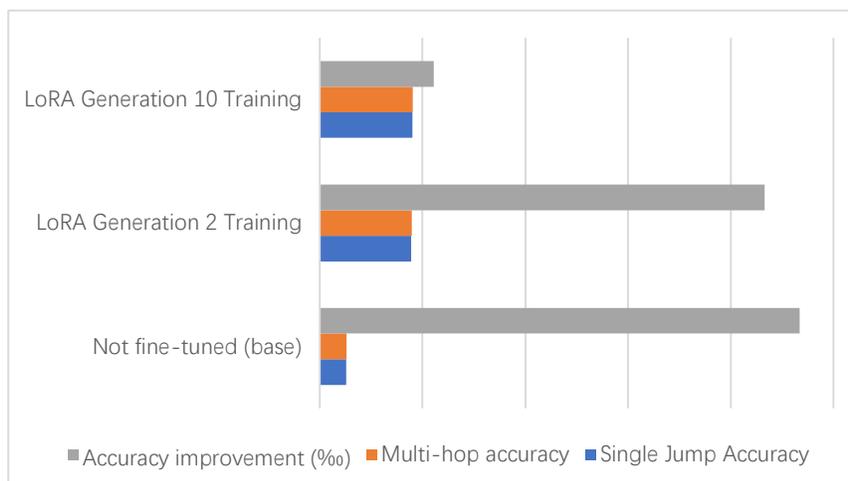

**Figure 1**. Model Performance Visualization

From the experimental results, we observe that the multi-hop decomposition strategy consistently outperforms the single-hop baseline across all configurations. In the non-fine-tuned scenario, although both accuracies are low, the multi-hop method shows a slight advantage (25.93% vs. 25.47%), indicating its benefit even without training. After applying LoRA fine-tuning, the accuracy improves significantly for both methods. At epoch 2, multi-hop accuracy reaches 89.32%, slightly higher than single-hop at 88.89%. By epoch 10, the difference narrows further (90.44% vs. 90.33%).

These results suggest that multi-hop decomposition enhances logical structuring and model understanding, especially in early training stages. Although the gap diminishes with more fine-tuning, the strategy still retains a consistent edge. This validates the method's generalization ability and practical value, particularly in zero-shot or few-shot inference scenarios.

## 5. Conclusion

This study aims to address the challenge of accurately answering complex questions using Large Language Models (LLMs) by introducing a multi-hop question decomposition method. Building on the MQUAKE framework and leveraging the LLAMA3 model, the research investigates how breaking down complex queries into sub-questions impacts model reasoning. The primary objective is to enhance inference accuracy through structured data preprocessing and fine-tuning techniques.

Through data analysis, we identified (1) multi-hop decomposition improves accuracy even without model fine-tuning, (2) fine-tuning significantly boosts overall performance, and (3) multi-hop decomposition maintains an advantage across all training stages. These findings suggest that structured question decomposition enhances the model's logical reasoning and generalization capabilities.

The results of this study have significant implications for the field of natural language processing. Firstly, the benefit of multi-hop decomposition provides a new perspective on reasoning enhancement in LLMs. Secondly, the consistent performance improvement challenges the idea that complex training alone is sufficient for high performance. Finally, the approach opens new avenues for combining decomposition strategies with real-time knowledge updates and minimal-resource training environments.

Despite the important findings, this study has limitations, such as a modest performance gap in high-accuracy settings and reliance on manually designed decomposition. Future research could explore automatic decomposition generation and extend the approach to multimodal or domain-specific datasets.

In conclusion, this study, through structured data preprocessing and LoRA-based fine-tuning of LLAMA3, reveals the effectiveness of multi-hop question decomposition in improving complex question answering. It provides new insights for enhancing reasoning capabilities in large language models.